\documentclass[letterpaper, 10 pt, final, conference]{ieeeconf}  

\IEEEoverridecommandlockouts                              

\usepackage{graphics} 
\usepackage{graphicx}
\usepackage[font=footnotesize]{caption}
\usepackage{subcaption}
\usepackage{amsmath} 

\RequirePackage{multirow} 
\RequirePackage{multicol}
\RequirePackage{array} 
\RequirePackage{booktabs} 

\title{\LARGE \bf
Leveraging distributed contact force measurements for slip detection: \\a physics-based approach enabled by a data-driven tactile sensor
}

\author{Pietro Griffa, Carmelo Sferrazza and Raffaello D'Andrea
	\thanks{The authors are members of the Institute for Dynamic Systems and Control, ETH Zurich, Switerland. Email correspondence to Carmelo Sferrazza
		{\tt\small csferrazza@ethz.ch}}%
}

\begin{document}

\maketitle
\thispagestyle{empty}
\pagestyle{empty}

\begin{abstract}


Grasping objects whose physical properties are unknown is still a great challenge in robotics. 
Most solutions rely entirely on visual data to plan the best grasping strategy.
However, to match human abilities and be able to reliably pick and hold unknown objects, the integration of an artificial sense of touch in robotic systems is pivotal.
This paper describes a novel model-based slip detection pipeline that can predict possibly failing grasps in real-time and signal a necessary increase in grip force.
As such, the slip detector does not rely on manually collected data, but exploits physics to generalize across different tasks. To evaluate the approach, a state-of-the-art vision-based tactile sensor that accurately estimates distributed forces was integrated into a grasping setup composed of a six degrees-of-freedom cobot and a two-finger gripper. 
Results show that the system can reliably predict slip while manipulating objects of different shapes, materials, and weights. 
The sensor can detect both translational and rotational slip in various scenarios, making it suitable to improve the stability of a grasp.

\end{abstract}


\section{INTRODUCTION} 
\label{sec:introduction}

In grasping tasks, the feedback provided by the sense of touch is paramount, as proved in humans.
Humans can efficiently manipulate a vast number of different objects, including even previously unknown ones, by regulating the grip force exerted. 
While at the beginning of the grasp gripping force is generated depending on the estimation of the load, this is then altered based on tactile feedback. 
This behaviour, which is dominated by the spinal cord for humans, is recognized as reflex control \cite{reflex_control}. In most of the cases, slippage of an object is prevented by humans with a safety margin of 10-40\% above the required minimum grasping force \cite{johansson2007tactile}.

A human level of dexterity in manipulation tasks is still unmatched in autonomous systems \cite{artificial_sense_slip}.
To date, the predominant input modalities explored in the robotic grasping literature are vision and depth. 
However, relying on these as the only sensory inputs for such applications is limiting, as it implies a delay in the reaction to a displacement caused by slippage.
In this regard, touch sensing has proven to be of great help towards more successful and consistent robotic grasping \cite{calandra2017feeling}, by instead providing information about the ongoing contact forces.
In particular, this work focuses on predicting slippage by monitoring the three-dimensional forces acting on the manipulated object, which is unknown to the algorithm.
This task represents the first step towards  replicating a human-like reflexive behaviour, aiming to correct grip forces once slippage is detected, and enabling fine balancing of the contact forces and minimum-force grasping. 

\begin{figure}[t]
  \centering
  \includegraphics[width=0.88\columnwidth]{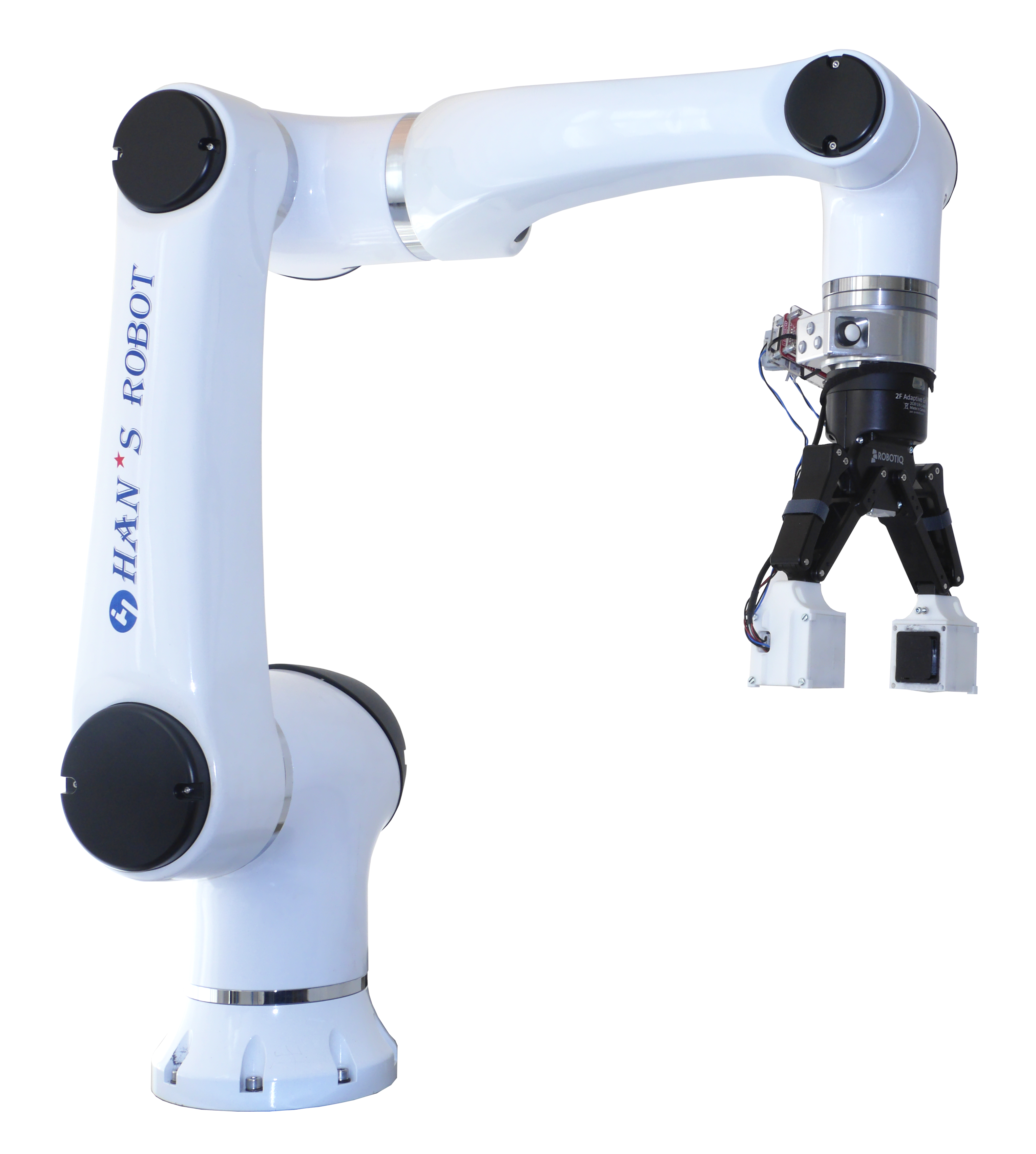}
  \caption{The setup employed for the experiments presented in this paper, comprising robot arm, gripper, and tactile sensor.}
  \label{fig:arm}
\end{figure}

\subsection{Related work}
\label{sec:related_work}

Research on artificial tactile sensing dates back to the 1970s \cite{tactile_sensors_for_friction_estimation_review} and has been steadily progressing over the years, with the ultimate objective of matching human manipulation capabilities, exploited in about 60\% of daily life activities \cite{FARIA2012396}.
Many different types of transduction methods have been proposed in the literature for robotic tactile sensing, the most common examples being resistive, capacitive, optical, piezoelectric, and magnetic-based sensors \cite{yousef2011tactile, nicholls1989survey}.
All  these devices aim to estimate contact-related quantities, but do not generally return distributed quantities. 
In particular, they mostly focus on measuring the force acting at a single point or the averaged force in the measurement area, with low spatial resolution. 
However, as pointed out in \cite{review_tactile_image_sensors, wan2018modeling}, spatial resolution and the number of tactile sensing elements (commonly referred to as ``taxels''), are critical factors for sophisticated robotic tactile sensing.


In recent years, there has been great and increasing interest in vision-based tactile sensors, mostly guided by the dramatic progress of image processing hardware and computer vision software, in terms of both performance and accessibility. 
These sensors, which are also referred as optical tactile sensors, work by converting physical stimuli to a form of distributed light that can be captured by a camera. 
The pixel number of modern cameras renders high spatial resolution, potentially beyond the tactile sensing capabilities of humans.
Vision-based tactile sensors are usually composed of three components: a tactile skin with physical contact-light conversion medium, a camera, and a computer. 
A broad overview of this type of device was given in \cite{review_tactile_image_sensors}, which identified two main subcategories based on the method of conversion from physical contact to light signal. 
Sensors based on a reflective membrane \cite{GelSight,DIGIT} generally track tactile imprints captured by the camera, while another subcategory tracks the displacement of markers distributed within a deformable surface \cite{sferrazza2019design,ward2018tactip,yamaguchi2016combining}. 
In general, vision-based tactile sensors benefit from high resolution, low costs, and ease of manufacture and wiring, despite usually being thicker than the other categories. 
While most vision-based tactile sensors in the literature focus on estimating total forces, contact locations, or surface deformation, \cite{sferrazza2020simtoreal} describes a technique to accurately estimate the three-dimensional contact force distribution in real-time.

The rich feedback provided by tactile sensors can be exploited to improve the performance of robots in grasping tasks, as shown in \cite{calandra2017feeling} for the prediction of grasp success, and in \cite{Calandra_2018} for grasp correction. 
Extensive overviews of how tactile sensors have been employed to recreate an artificial sense of slip were presented in \cite{artificial_sense_slip, tactile_sensors_for_friction_estimation_review}.
However, as pointed out in \cite{artificial_sense_slip}, the human sense of slip is hard to emulate, because there is no single specific receptor devoted to this task that can be mimicked.
This lack of a biological model to provide inspiration has resulted in a wide variety of approaches being proposed in the literature, differing from each other in terms of parameters to be monitored, the transduction mechanism, and technological solutions.
In particular, force-based methods guarantee a good trade-off between complexity and robustness, as described in \cite{artificial_sense_slip}. 
These are often based on monitoring the resulting (also referred to as total) normal force needed to balance the resulting tangential force, assuming a known friction coefficient. 
However, since total forces do not carry information about the traction field, these methods generally struggle to capture rotational slip.

On the other hand, the vast majority of works based on distributed measurements (e.g., pressure sensor arrays, or vision-based tactile sensors) to detect slip have focused on data-driven approaches. 
For example, in \cite{assessinggrasp_stability}, discrete pressure maps provided by tactile sensors were employed to predict grasp stability. 
In contrast, the solution proposed in \cite{9216788} leveraged a vision-based tactile sensor and the properties of recurrent neural networks to predict the variation trend of the frictional force at the moment of contact, and classify the contact state. Data-driven methods generally provide computationally efficient inference, which may be critical for detecting slippage in time, and overcome the need to model contact.
However, these approaches are very sensitive to training data, which are still hard to collect consistently. 
In fact, to date, the most common way of labeling slip data is by doing it manually (sometimes referred as ``expert labeling'' \cite{zhang2019learning, 9216788}), relying on human intuition. 
This process is generally time-consuming for large datasets and often results in imprecise labels, due to errors introduced by the operator.
While some works have tried to eliminate the human in the loop by using vision to label the slip data \cite{calandra2017feeling}, these methods struggle to recognize smaller displacements, and are more often employed to detect whenever the grasped object is completely dropped.

To bypass the dependency on training data while retaining reliable prediction accuracy about the slip state, this work presents a model-based approach to slip detection. 
Compared to other force-based methods, here distributed contact force measurements are leveraged to measure the slipping area in general scenarios, including both translational and rotational slip. 
The use of physics to monitor slip phenomena exploits the precise tactile feedback provided by the sensing technique described in \cite{sferrazza2020simtoreal}, and efficiently enables accurate slip predictions.

\subsection{Outline}
\label{sec:outline}

The sensing principle and the sensor used for the experiments are discussed in Section \ref{sec:sensing_principle}.
Section \ref{sec:method} describes the approach to slip detection.
The experimental setup and the related results are discussed in Section \ref{sec:experiment}, while Section \ref{sec:conclusion} draws conclusions and gives an outlook on future work.
\section{TACTILE SENSOR} 
\label{sec:sensing_principle}

\begin{figure}[t]
  \centering
  \includegraphics[width=0.98\columnwidth]{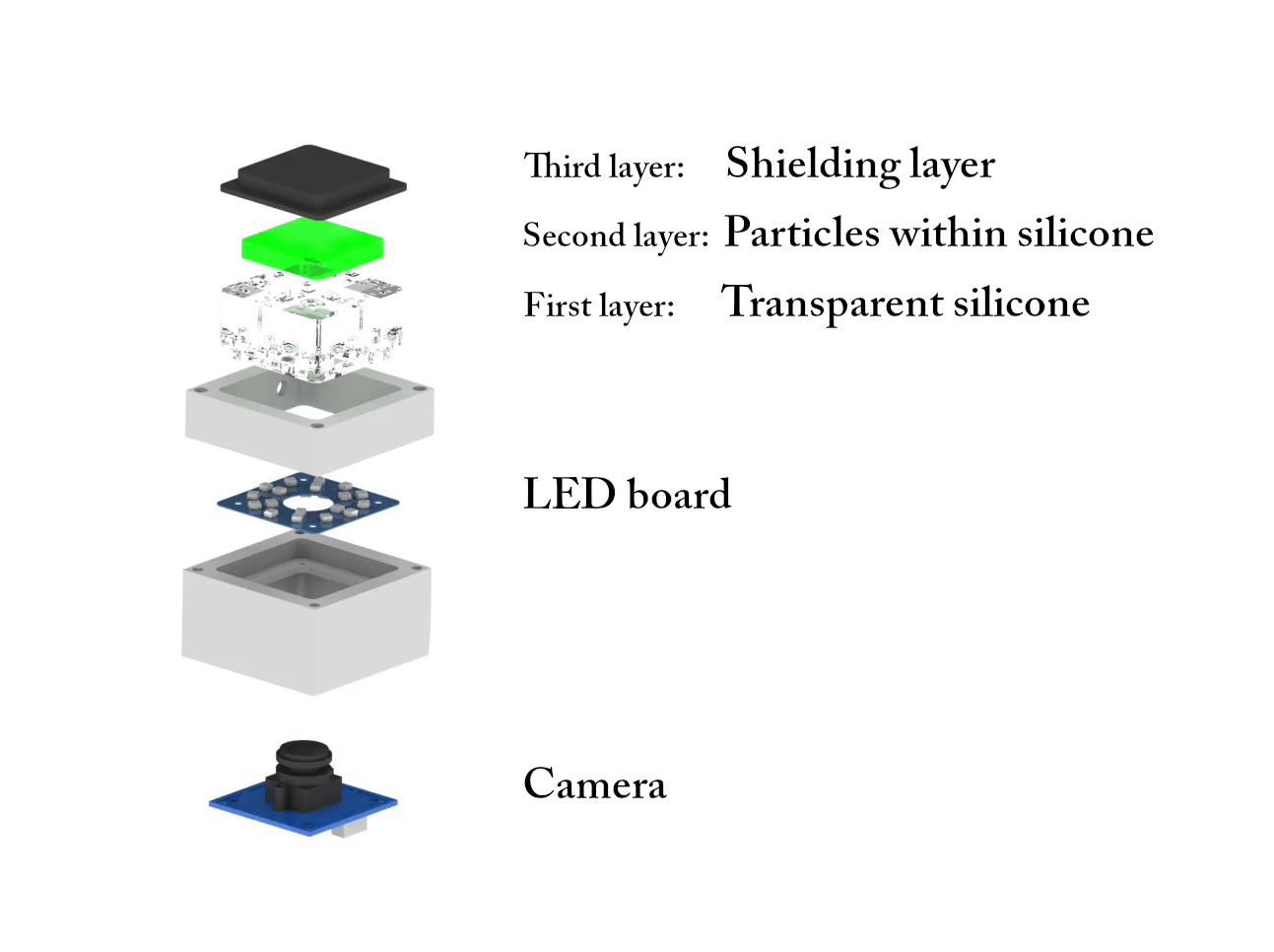}
  \caption{Exploded view of the tactile sensor.}
  \label{fig:sensor}
\end{figure}

The vision-based tactile sensor employed in this work is based on the principle described in \cite{sferrazza2019design}, mounting an internal camera that tracks particles randomly distributed within a deformable material.
Such soft silicone gel is produced in a three-layer structure, resting on top of a camera surrounded by LEDs that guarantee uniform brightness.
The layers, cast in sequence (see Fig.~\ref{fig:sensor}), are, from the inside outwards:
\begin{itemize}
    \item A stiff transparent layer (ELASTOSIL RT 601 RTV-2, mixing ratio 7:1, shore hardness 45A), which serves as a spacer and for light diffusion;
    \item A transparent layer (ELASTOSIL RT 601 RTV-2, mixing ratio 25:1, shore hardness 10A), with spherical green polyethylene particles (microspheres with a diameter of 150 to 180 $\mu m$) spread within its volume;
    \item A black layer (ELASTOSIL RT 601 RTV-2, mixing ratio 25:1, shore hardness 10A), which shields the sensor from external light. 
\end{itemize}
The two outermost layers have both been previously characterized \cite{8918082} as hyperelastic materials using second-order Ogden models \cite{doi:10.1098/rspa.1972.0096}, while, due to the large difference in stiffness, the stiff base layer was modeled as rigid.
The camera mounted is an ELP USBFHD06H, equipped with a fisheye lens that has a field of view of 180 degrees.
The images were captured at a resolution of $640\!\times\!480$ pixels, and subsequently cropped to $440\!\times\!440$ pixels.

\begin{figure}
\begin{subfigure}{.98\columnwidth}
  \centering
  \includegraphics[width=.96\columnwidth]{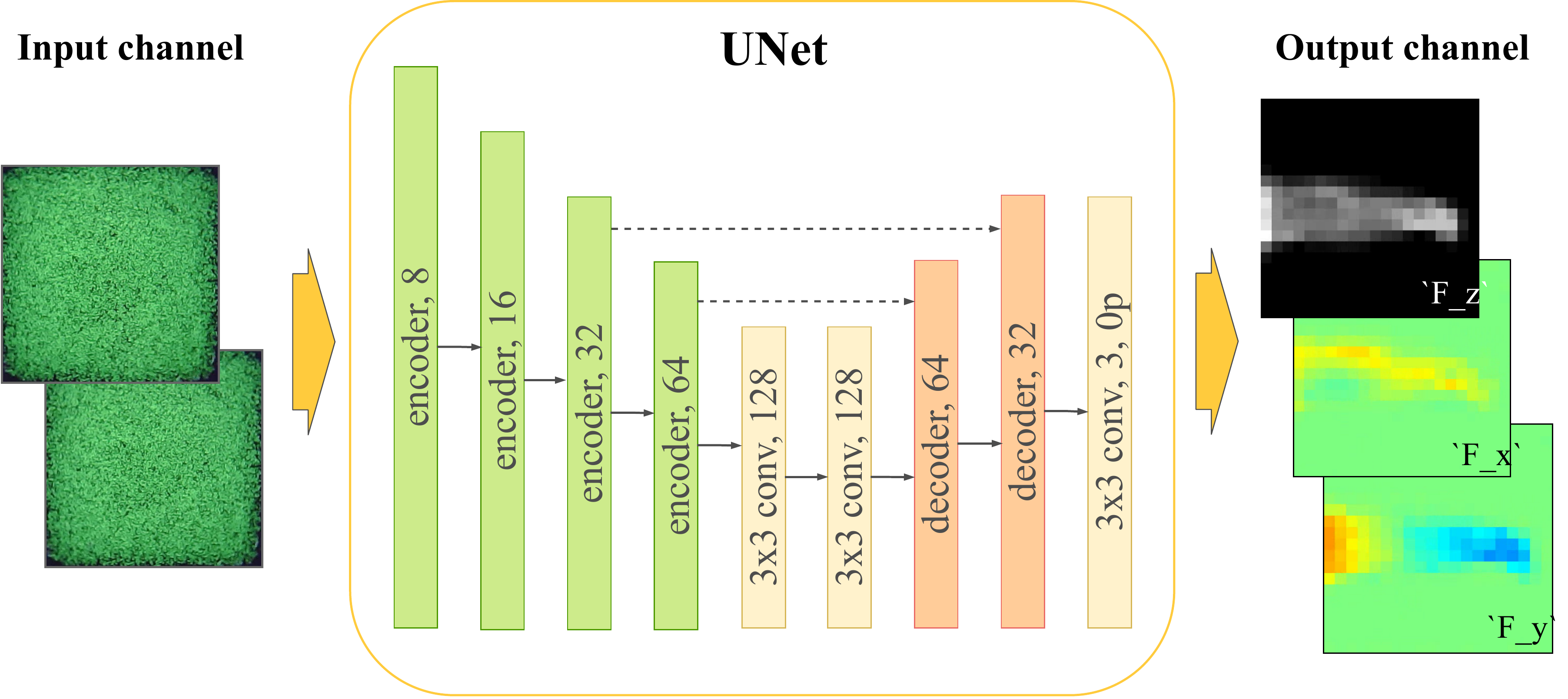}  
  \caption{Learning architecture.}
  \label{fig:sub-first}
\end{subfigure}
\newline
\begin{subfigure}{.48\columnwidth}
  \vspace{5pt}
  \centering
  \includegraphics[width=.47\columnwidth]{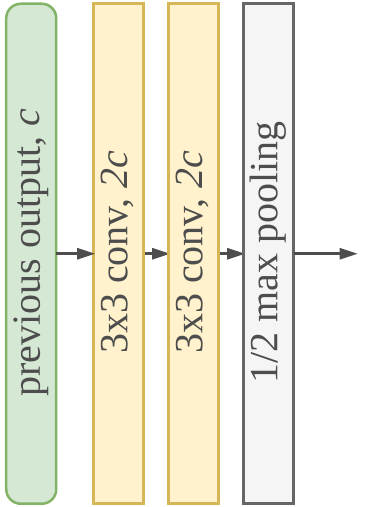}  
  \caption{An encoder block.}
  \label{fig:sub-second}
\end{subfigure}
\begin{subfigure}{.48\columnwidth}
  \vspace{5pt}
  \centering
  \includegraphics[width=.47\columnwidth]{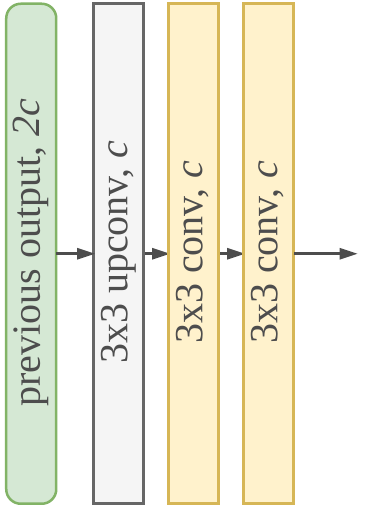}  
  \caption{A decoder block.}
  \label{fig:sub-third}
\end{subfigure}
\caption{In (a), a diagram of the learning architecture is shown. The neural network outputs the three components of the force distribution, represented above as images, with the corresponding color bars omitted for ease of visualization. The encoder and decoder blocks are summarized in (b) and (c), respectively. ``$3\!\times\!3$ conv, $c$'' indicates a convolutional layer with a $3\!\times\!3$ filter size and $c$ output channels, while ``$3\!\times\!3$ upconv, $c$'' indicates an upconvolution that doubles the input size. The dotted lines indicate the concatenation of an earlier layer output with upsampled information. After each convolutional layer, with the exception of the one before the final output, batch normalization and rectified linear units were employed. ``0p'' indicates no padding. Where not indicated, all convolutional filters have unit zero-padding and unit stride.}
\label{fig:dnn}
\end{figure}

Camera frames are fed to a fully convolutional deep neural network (DNN), shown in Fig.~\ref{fig:dnn}, that returns the discretized force distribution applied on the surface of the tactile sensor.
The learning task is addressed as a pixel-wise regression, with inputs in the form of two $88\!\times\!88$ images: a zero image taken at rest and the most recent frame. 
The DNN output represents the 3D force distribution as a $3\!\times\!n\!\times\!n$ image-like tensor, where $n\!\times\!n$ is the number of taxels the sensor's surface is discretized into.
In the context of this work, $n$ was fixed to 20, for a total of 400 measurement points.
The first two components of this distribution represent the $x$ and $y$ components of shear force, laying on the plane of the sensor's surface at rest, while the third component contains the value of the force acting in the direction $z$ normal to the surface.
\\
All the training data were generated entirely in simulation, following the strategy proposed in \cite{sferrazza2020simtoreal}.
A number of indentations were simulated on the sensor via the finite element method (FEM), generating pairs of synthetic tactile images and associated ground truth labels, representing the 3D contact force distribution applied to the sensor's surface.
The DNN is finally trained on a dataset consisting of about 95,000 datapoints, obtained with seven different indenters' geometries.
Moreover, at training time, the dataset is randomly augmented by appropriately flipping the features and labels, exploiting the symmetry of the gel geometry and the camera projection. 
Additionally, in order to address the sim-to-real gap, the images were perturbed with random salt-and-pepper noise and brightness changes.

\begin{figure}[h]
  \centering
  \includegraphics[width=0.98\columnwidth]{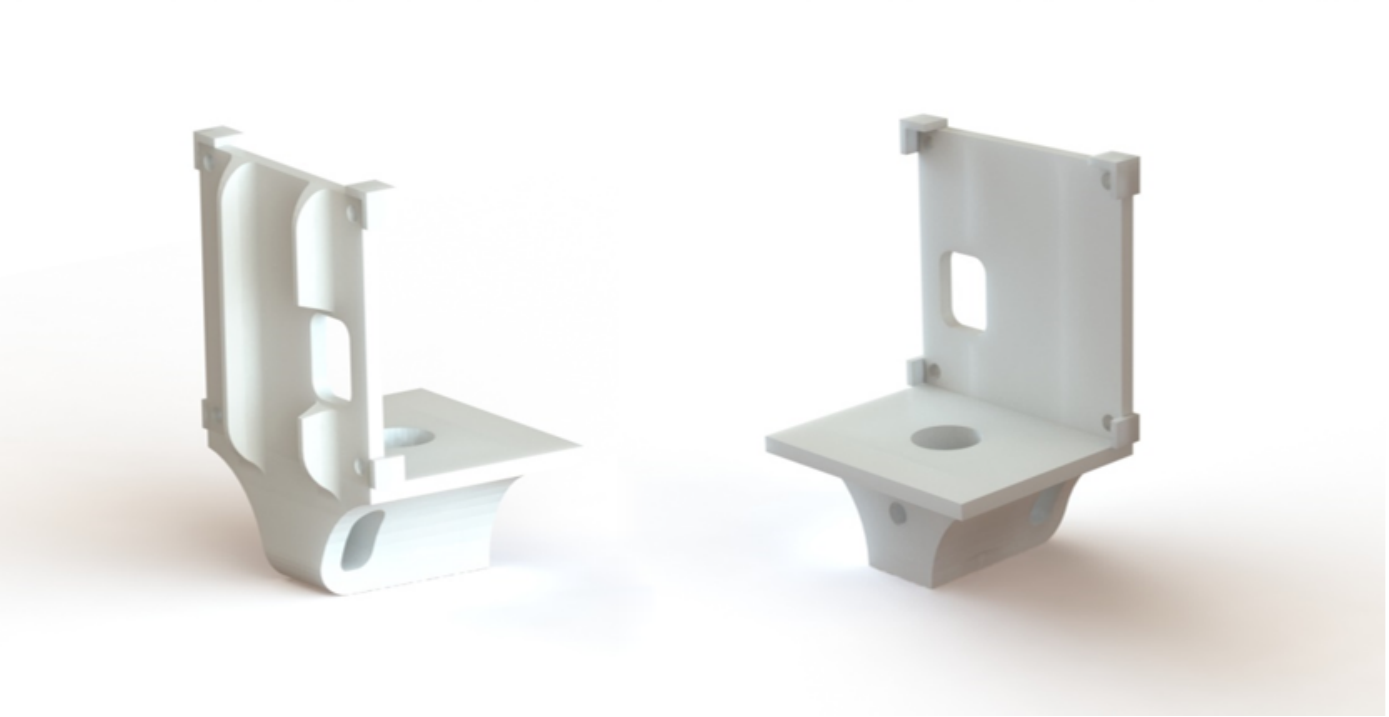}
  \caption{Custom housing to accommodate the tactile sensor. After securing the housing to the gripper with a screw through the hole in the bottom part of the component, the sensor can be fixed to the housing through four M3 screws. The opening in the back plate of the housing enables an effective cable management. Reinforcements in the back of the component serve to increase its bending strength.}
  \label{fig:ts_housing}
\end{figure}

In order to seamlessly integrate the tactile sensor into a complete grasping setup, composed of a manipulator and gripper, such as shown in Fig. \ref{fig:arm}, a custom housing was designed, see Fig.~\ref{fig:ts_housing}.
Manufactured using HP PA12 through jet fusion, the housing ensures straightforward assembly to the gripper, requiring a single screw and two dowel pins that serve to maintain alignment and prevent rotation. 
The gripper described in Section \ref{sec:experiment} was equipped with two such tactile sensors in place of the fingertips. 
However, for the sake of computational efficiency, only the readings from one of the tactile sensors were considered in this work.
This was made possible by the focus of this research on motion along the surface of the tactile sensor and rotation along its normal axis. 

\section{METHOD} 
\label{sec:method}

In the following section, the proposed approach to slip detection is discussed.
First, a baseline method based on monitoring total forces is introduced. Then, this baseline is compared to the novel solution proposed.

\subsection{Baseline} 
\label{sec:baseline}

A stable grip requires the compression force to be strong enough to generate a friction force sufficient to balance the weight of the manipulated object, in addition to possible external loads. 
It is in fact known that, in general, the friction force increases proportionally to the grasping force, and depends just on the materials of the slider and substrate (the grasped object and the fingertip in the case considered here), information often encapsulated in the form of a friction coefficient.
As remarked in \cite{artificial_sense_slip}, slip events may occur if at least one of the following conditions is verified:
\begin{enumerate}
    \item the normal grip force is insufficient; 
    \item the tangential force is higher than expected or quickly increasing;
    \item the friction coefficient becomes too low, as in the case of wet and slippery surfaces.
\end{enumerate}
In the context of this work, only dry friction was considered: as such, the focus is entirely on monitoring shear and normal forces, to modulate accordingly the normal grip force whenever it is estimated not to balance the tangential force properly. 
The most widespread dry friction law is undoubtedly the one named after Coulomb-Amontons-DaVinci (from here on referred as the Coulomb friction law for simplicity). 
This friction model asserts that in sliding systems consisting of a solid object on a solid substrate under dry conditions, the frictional force is proportional to the loading force and does not depend on the apparent contact area, net displacement, or slip history.
It results that during slip the friction force at a point is opposite to the direction of motion, and its magnitude is not influenced by it. 
According to this model, slip happens when the external load exceeds the friction force in magnitude, which depends uniquely on the loading force and the friction coefficient. The slip condition then reduces to:
\begin{equation}
    F_T > \mu F_N,
    \label{eq:coulomblaw}
\end{equation}
where $\mu$ is the friction coefficient, and $F_N$ and $F_T$ are the magnitude of the total normal and shear force components of the contact force. \\

As described in Section \ref{sec:sensing_principle}, the tactile sensor employed in this work returns tactile data in the form of three-dimensional matrices containing normal and shear forces per taxel.
From this data, it is possible to compute total quantities as the integrals over the entire contact region of the force distribution. Receiving tactile data as a series of discrete contact points, the integrals are simplified as summations.
Therefore the magnitude of the normal force $F_N$ acting on the sensor can be computed as
\begin{equation}
    F_N =  \sum_{i=0}^{{n^2}-1} f_{z_i},
    \label{eq:totalNormalForce}
\end{equation}
where $f_{z_i}$ refers to the vertical component (along the $z$ axis) of the measured force, and the subscript $i$ refers to measurements relative to the $i$-th taxel.  \\
Similarly, the horizontal components of the total contact force can be computed as
\begin{equation}
    F_x =  \sum_{i=0}^{{n^2}-1} f_{x_i}
    \label{eq:totalShearX}
\end{equation}
\begin{equation}
    F_y =  \sum_{i=0}^{{n^2}-1} f_{y_i},
    \label{eq:totalShearY}
\end{equation}
where $f_{x_i}$ and $f_{y_i}$ are the $x$ and $y$ components, respectively, of the measured force at the $i$-th taxel. From such quantities, the total shear force can be easily computed as
\begin{equation}
    F_T =  \sqrt{F_x^2 + F_y^2}.
    \label{eq:totalShearForce}
\end{equation}
Note that also the net moment about the $z$ axis may be computed as 
\begin{equation}
    M =  \sum_{i=0}^{{n^2}-1} (x_i f_{y_i} - y_i f_{x_i}),
    \label{eq:totalMoment}
\end{equation}
where $[x_i, y_i]$ are the coordinates of the $i$-th taxel with respect to the origin. \\
Therefore, $F_T$ and $F_N$ may be employed to monitor slip according to \eqref{eq:coulomblaw}, assuming a known friction coefficient. \\

Even though the Coulomb friction law provides a straightforward condition to detect gross slip, a term that refers to when all parts of both contact surfaces slip completely against each other, it has some major limitations.
The most relevant for this application is that it is defined for contact between rigid bodies, and only where the problem can be reduced to point contact, thus not providing any measure of rotational slip. 
In the case of soft contact, the contact patch assumes a finite area, and the assumption of point contact is no longer applicable, thus requiring a different approach to the problem. 

\subsection{Stick ratio approach} 
\label{sec:ours}

To address the limitations of the Coulomb friction law, more elaborate friction models have been proposed in the literature over the years \cite{GOYAL1991307, modeling_contact_for_soft_fingers}, in order to expand the formulation to be able to handle rotational slip and soft contact.
However, these models are often characterized by a computational complexity that makes them unsuitable for real-time applications on resource-limited platforms. 
For this reason, an alternative approach to the problem is proposed in this work, based on the concept of incipient slip, which refers to when relative displacement occurs only in a narrow region of the contact interface, and always precedes gross slip. 
In fact, it can be observed that slip commonly starts from just a portion of the contact surface, followed by an expansion of the the slip area with the increase of the applied tangential force. 
Finally, when the slip area expands to the whole contact area, gross slip occurs. 

Despite the limitations discussed in the previous subsection, the Coulomb friction law is valid \textit{locally} for elastic surfaces under dry conditions \cite{Otsuki_2013}. 
Therefore, the discretized fashion in which the tactile sensor employed returns the contact forces is leveraged here to detect at each timestep the taxels inside the contact area that are contributing to slippage. 
This is done by considering each of the taxels as a discrete contact point and comparing normal and shear forces at the single point.
The ratio of stick area ($C$) against contact area ($A$) is commonly called stick ratio ($SR$), and represents a measure of incipient slip, thus enabling gross slip prediction.
A formal definition is the following:
\begin{equation}
    SR = \frac{C}{A} = \frac{|\{i: f_{T_i} \leq \mu f_{z_i}\}|}{|\{i: f_{z_i} > 0\}|},
    \label{eq:srdefinition}
\end{equation}
where $f_{T_i}:=\sqrt{f_{x_i}^2+f_{y_i}^2}$ describes the shear force at the $i$-th taxel, with $C$ and $A$ computed through the cardinality of the sets of taxels sticking and in contact, respectively. 
Note that the constant size of the taxels cancels out in this ratio.

As showed in \cite{theoretical_derivation_and_realization_of_adaptive_grasping}, slip is commonly anticipated by a sudden drop of the stick ratio. 
Following this reasoning, sudden drops must be avoided for stable grip.
As such, a straightforward way of identifying slip is provided by monitoring the trend of the value of $SR$, as computed in equation \eqref{eq:srdefinition}.
\section{EXPERIMENT} 
\label{sec:experiment}

In this section, the robot system employed to evaluate the proposed slip detection algorithm is introduced, demonstrating how the experiments were conducted and discussing their results. \\
The performance of the proposed strategy is compared to a baseline provided by a straightforward implementation of the Coulomb friction law on the forces read by the tactile sensor.
Both approaches require knowledge of the friction coefficient's value, which in this case is assumed to be constant.
Indeed, experimental data showed that this assumption only leads to a limited loss of accuracy, due to the fact that the material of the tactile sensor surface is predominant with respect to most materials in sliding interactions with other objects, see \cite{sferrazza2020simtoreal}. 
In this work, the friction coefficient was tuned to a constant value of 0.45 for all experiments.

\subsection{Experimental setup}
\label{sec:experimental_setup}

The hardware setup consisted of the LARA 10 cobot by NEURA Robotics, used to realize consistent lifting trajectories, and a gripper model 2F-140 by Robotiq, housing two tactile sensors replacing the original fingertips.
The six degrees-of-freedom cobot and the two-fingered gripper were both controlled through an Intel NUC running Ubuntu 18.04, which received the necessary commands from the main machine.
All the central software components, including the contact force inference and the slip detection pipeline, were run on an Ubuntu 18.04 PC equipped with an Intel Core i7-3630Q @2.4GHz CPU. \\
Every experimental run was performed by running, for each test object, a grasping sequence defined a priori.
In general, the arm approached, grasped, and lifted the test object, and then an external load was manually applied until sliding happened between the tactile sensors and the grabbed object. 
For every frame captured by the sensor, the slip detection system predicted whether the object was either slipping (SLIP) or not (STICK).
During the whole process, the change of slip state, either STICK or SLIP, was recorded manually.
This information was then used in the form of ground truth labels to compute performance scores for every experiment. 
Note that similar approaches are generally employed to label data used to train learning-based slip detection strategies. 
In contrast, such labels were only used for evaluation in this work, where the algorithm is instead model-based.
\\
To test the effectiveness and reliability of the proposed approach in multiple scenarios this was tested on five objects, shown in Table \ref{tab:objs_res}, all of different shape, weight, and material.
For each object the procedure described earlier was performed several times, each time forcing five relative displacements, four of which were translational, and one rotational.
Specifically the sequence of motions was always according to the pattern T-T-R-T-T, where T stands for translational and R for rotational displacements.
To see an example of how such an experiment was conducted, refer to the attached video.

\begin{figure}[h]
  \centering
  \includegraphics[width=1 \columnwidth]{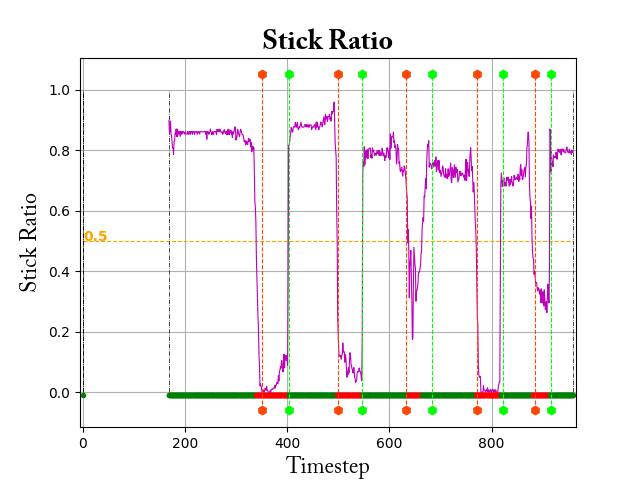}
  \caption{Stick ratio trend during a single experiment, comparing the proposed strategy's predictions (represented by the green or red circular markers, for STICK and SLIP respectively, along the $SR=0$ axis), against the ground truth data (whose change of state is identified by the colored vertical lines, with red ones representing the start of slippage, green ones the end of it).}
  \label{fig:stick_ratio}
\end{figure}

\subsection{Experimental results}
\label{sec:experimental_results}

\begin{table*}[t]
\begin{center}
 \begin{tabular}{
    | >{\centering\arraybackslash}m{0.075\textwidth}
    | >{\centering\arraybackslash}m{0.085\textwidth}
    | >{\centering\arraybackslash}m{0.11\textwidth}
    >{\centering\arraybackslash}m{0.11\textwidth}
    >{\centering\arraybackslash}m{0.11\textwidth}
    >{\centering\arraybackslash}m{0.11\textwidth}
    >{\centering\arraybackslash}m{0.11\textwidth} | }
 \hline
 \multicolumn{2}{|c|}{}
 & \vspace{4pt}\includegraphics[width=0.1\textwidth]{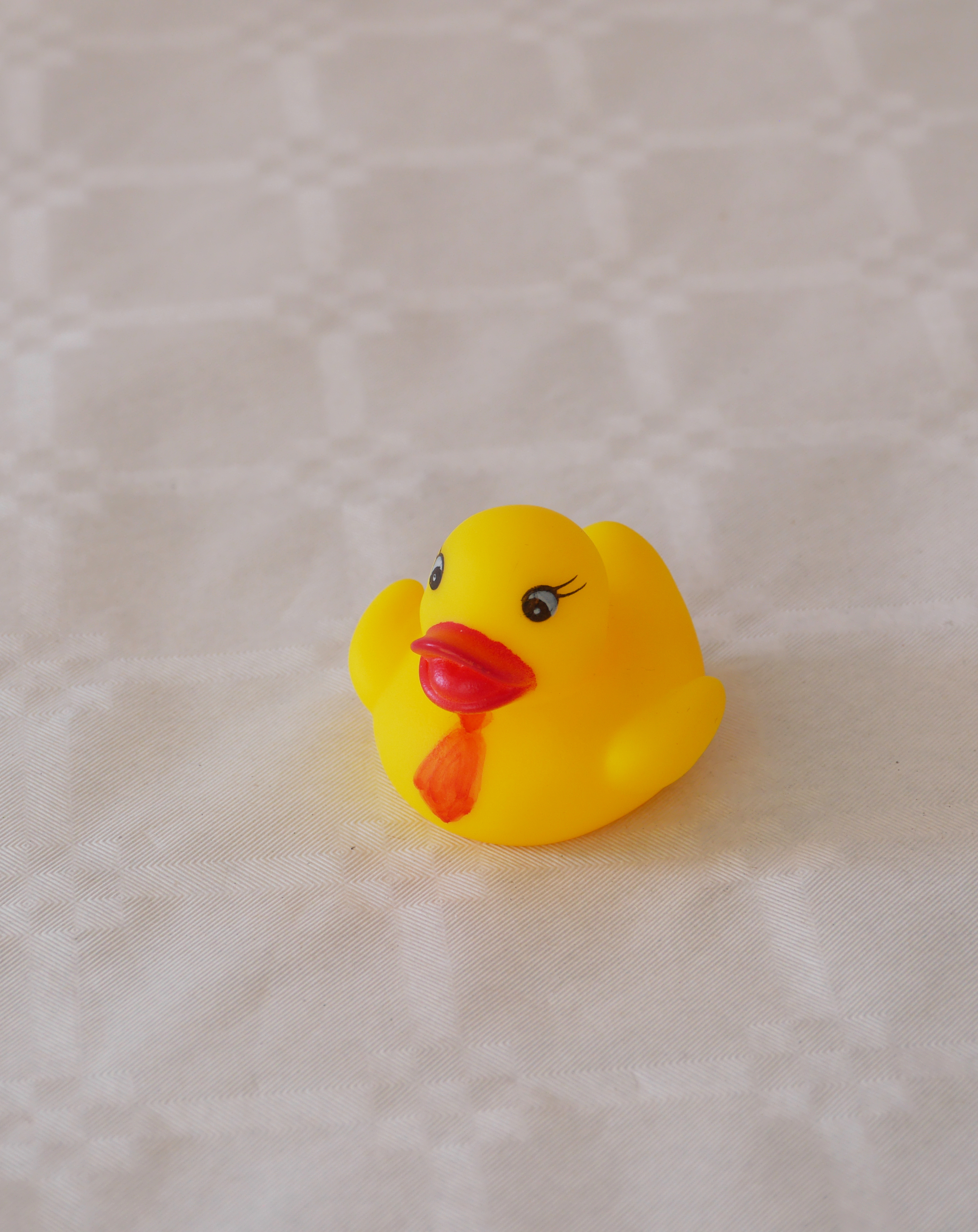} 
 & \vspace{4pt}\includegraphics[width=0.1\textwidth]{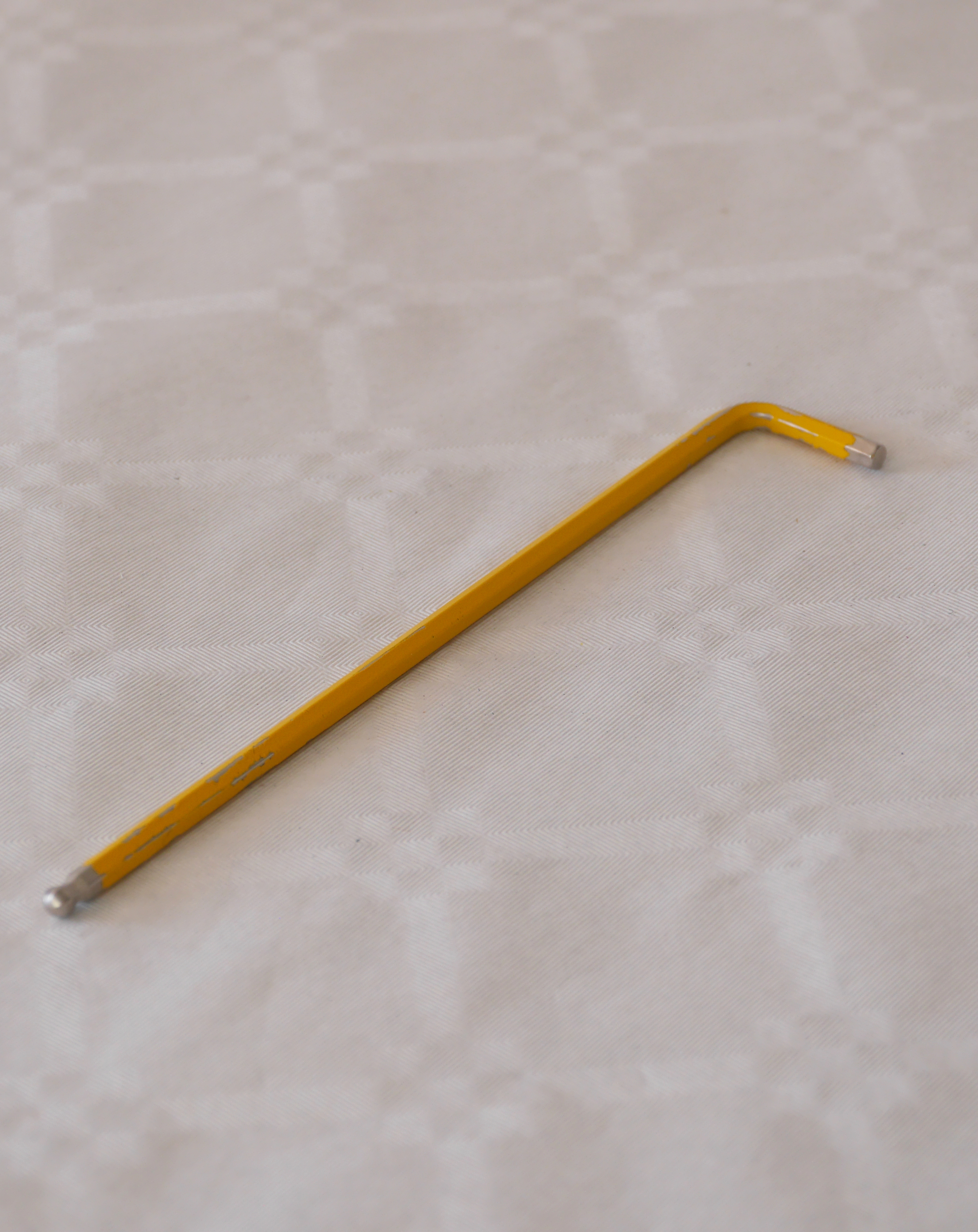} 
 & \vspace{4pt}\includegraphics[width=0.1\textwidth]{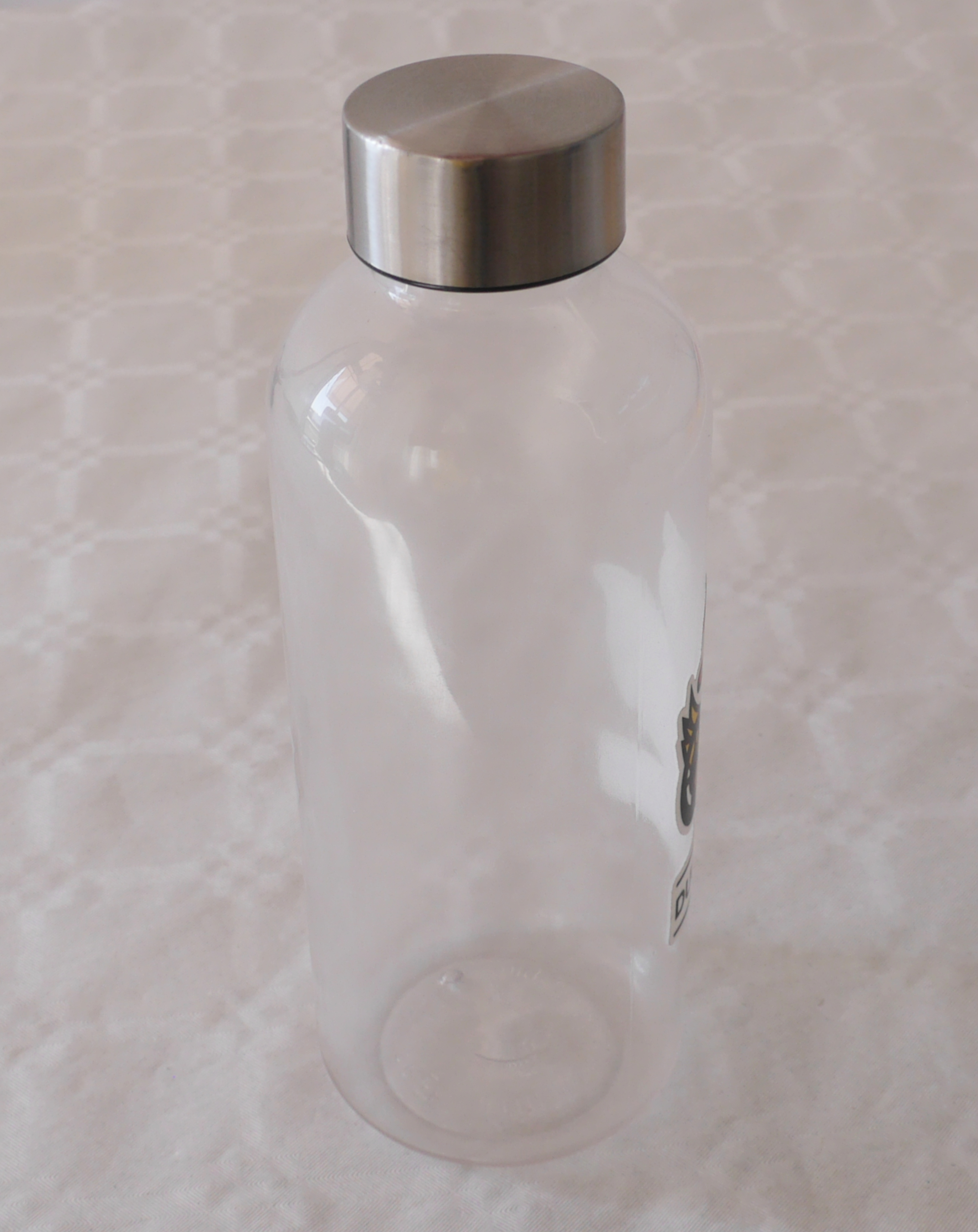} 
 & \vspace{4pt}\includegraphics[width=0.1\textwidth]{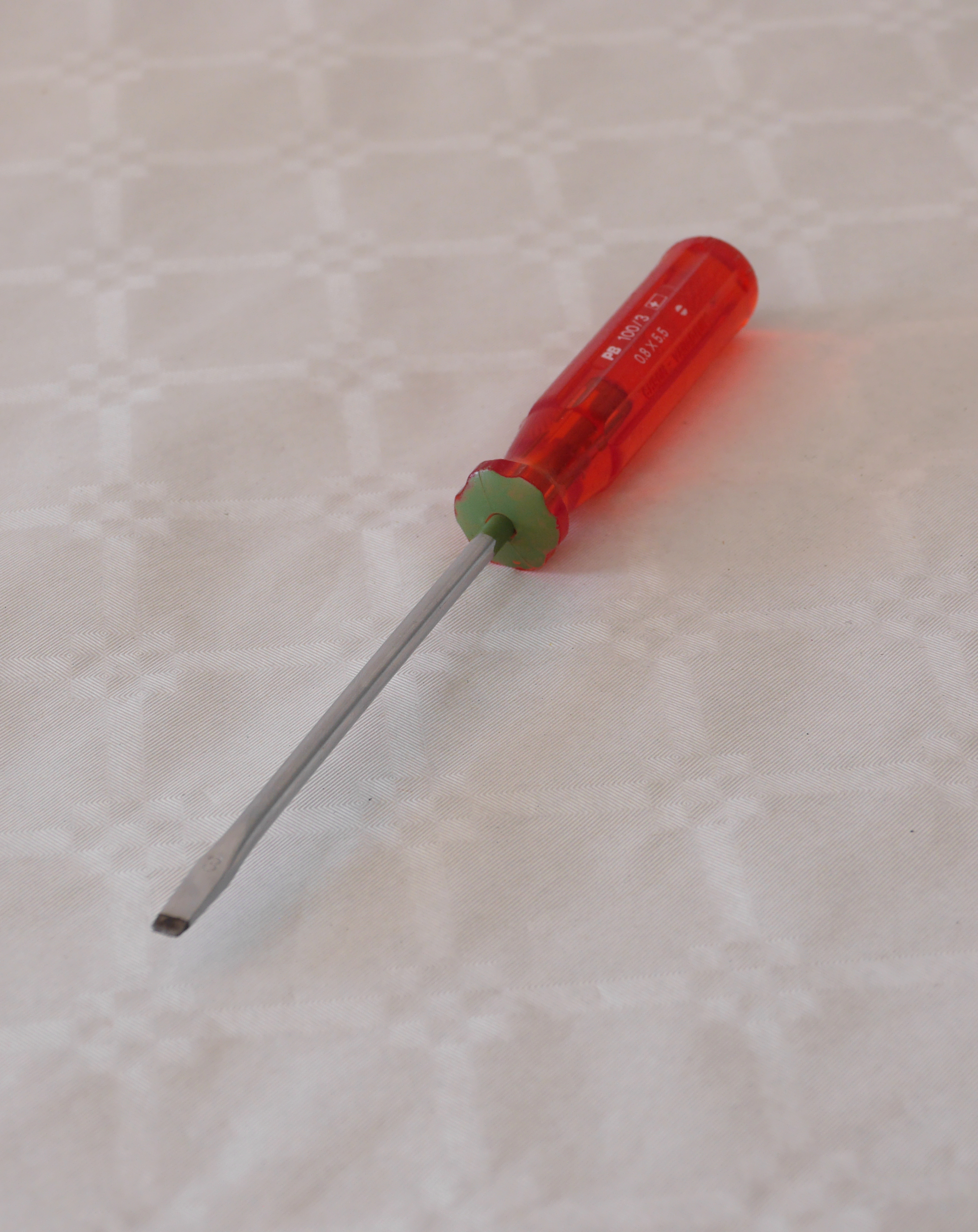} 
 & \vspace{4pt}\includegraphics[width=0.1\textwidth]{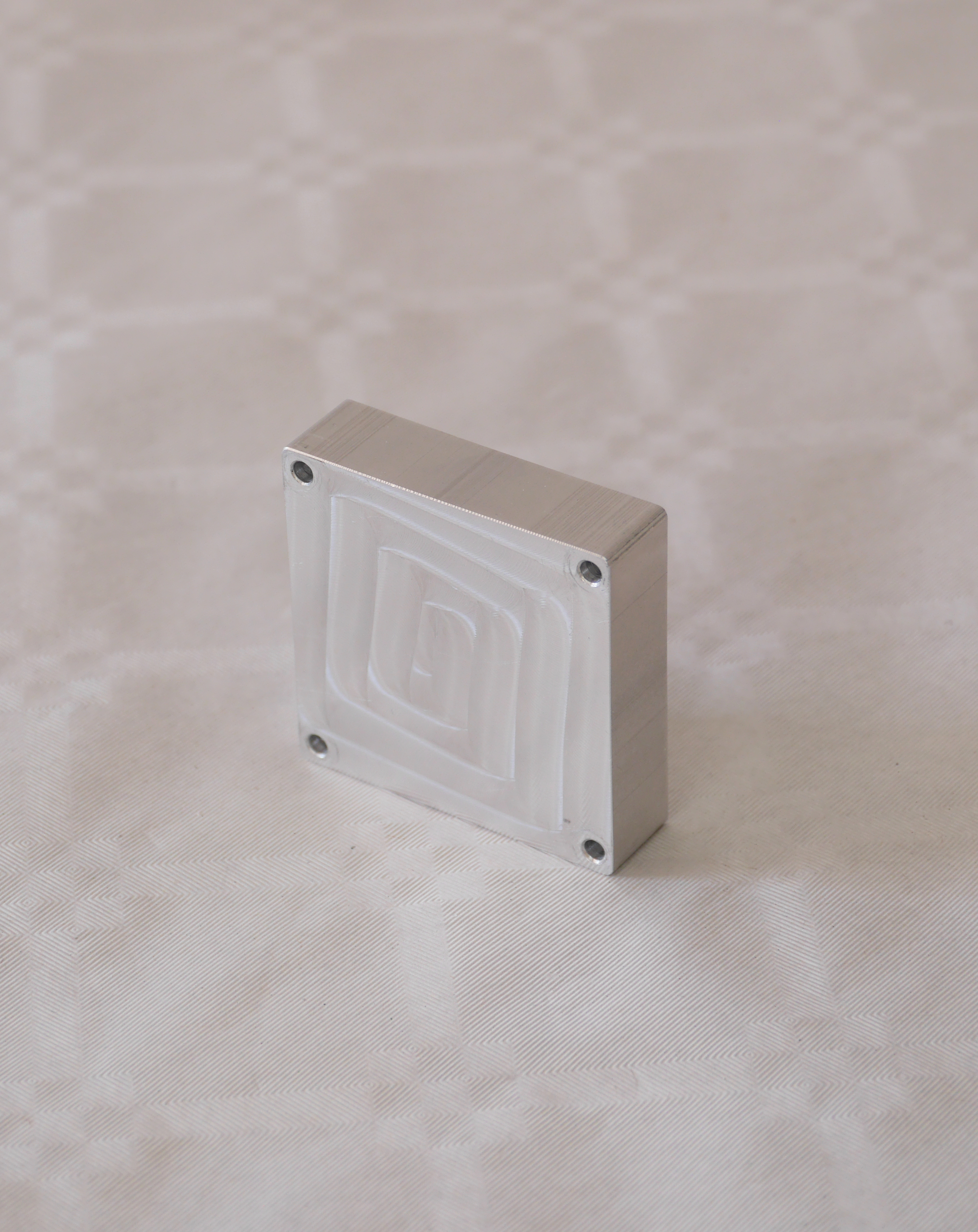}  \\
 \multicolumn{2}{|c|}{} & \footnotesize{11.90 g} & \footnotesize{17.82 g} & \footnotesize{91.11 g (empty)} & \footnotesize{57.50 g} & \footnotesize{92.66 g} \\
 \hline
 \multirow{3}{*}{Baseline} 
 & Accuracy & 79.01 \% & 74.40 \% & 75.71 \% & 78.12 \% & 76.10 \% \\
 & Precision & 64.19 \% & 83.02 \% & 76.00 \% & 79.61 \% & 60.83 \% \\
 & Recall & 61.60 \% & 62.93 \% & 57.45 \% & 65.17 \% & 63.20 \% \\
 \hline
 \multirow{3}{*}{Stick Ratio} 
 & Accuracy & 79.99 \% & 84.80 \% & 80.56 \% & 93.64 \% & 79.00 \% \\
 & Precision & 65.02 \% & 87.22 \% & 60.75 \% & 90.47 \% & 66.33 \% \\
 & Recall & 89.30 \% & 69.35 \% & 96.02 \% & 87.55 \% & 71.71 \% \\
 \hline
 \end{tabular}
 \caption{Scores per object.}
 \label{tab:objs_res}
\end{center}
\end{table*}

To evaluate and benchmark the performance of the method proposed, the evaluation metrics commonly used for binary classification were employed, namely accuracy, precision, and recall.
Particularly critical for the application considered is the recall, which represents a measure of how often the system predicts slip when there is actually a sliding motion.
To reduce the uncertainty resulting from the ground truth labels being assigned manually, all the results presented were obtained as the average of three different runs for each of the experiments. 
Table \ref{tab:objs_res} presents the resulting accuracy, precision and recall for both the baseline approach, applying the Coulomb friction law on total forces, and the proposed method based on monitoring the stick ratio.
The implementations of both these approaches run on Python on the laptop described in the previous subsection at a similar frequency, generally higher than 50 Hz, with small fluctuations depending on the size of the contact area.

The novel strategy to predict slip outperforms the baseline method, especially due to the ability to handle rotational slip in addition to translational slip.
The algorithm detects slippage whenever the stick ratio falls below an arbitrarily chosen threshold, which was set to 0.5 for all experiments presented in this section.
In general, this threshold can be used to tune the conservativeness of the method. 
Fig. \ref{fig:stick_ratio} shows the stick ratio trend during a typical experiment.
It is evident how slippage is generally accompanied by a sudden drop in the stick ratio value.
To better characterize the behavior of the proposed strategy, separate experiments were also performed by forcing one single type of relative displacement, either translational or rotational.
The results of such experiments are shown in Table \ref{tab:mot_res}.

\begin{table}[h]
\begin{center}
 \begin{tabular}{
    | >{\centering\arraybackslash}m{2.1cm}
    | >{\centering\arraybackslash}m{2.6cm}
    >{\centering\arraybackslash}m{2.6cm} | }
 \hline
 & Translational only
 & Rotational only  \\
 \hline
 \vspace{2pt}Accuracy & \vspace{2pt}85.98 \% & \vspace{2pt}86.86 \% \\
 Precision & 83.21 \% & 94.19 \% \\
 Recall & 85.65 \% & 79.51 \% \\
 \hline
 \end{tabular}
 \caption{Scores per sliding motion type for the proposed solution.}
 \label{tab:mot_res}
\end{center}
\end{table}
\section{CONCLUSION} 
\label{sec:conclusion}

In this paper, an approach to predict slip and signal possibly failing grasps has been discussed.
The proposed strategy leverages the features of a state-of-the-art tactile sensor to reliably monitor incipient slip in order to predict and avoid gross slip in grasping tasks.
In fact, here the sensor inference is decoupled from the force-based slip predictions. 
While the sensor's measurements are data-driven, its generalization capabilities \cite{sferrazza2020simtoreal} yield accurate force distribution in various scenarios. 
This enables reasoning at the level of forces, which are the essence of the accurate model-based approach to slip detection proposed here.
This pipeline provides sensible predictions for both translational and rotational motions, on a variety of objects and at high frequency, enabling real-time usage and potential online adjustment of failing grasps.

The remaining gap in performance observable between translational and rotational slip (see the recall row) can be explained by the marginal drop in accuracy of the contact forces returned by the tactile sensor during rotational motion.
This unwanted behaviour is mainly due to geometric interference between the slider and the edges of the squared tactile sensor when rotating the test object.
Such problems may be overcome by employing a dome-shaped tactile sensor similar to the one used in \cite{9444131}.
Future work will include integrating the proposed pipeline in grasp synthesis applications, in order to efficiently use tactile data alongside other sensory inputs, to refine the choice of grasp points, based on the slip predictions during the trajectory execution.  



\section*{ACKNOWLEDGMENT}

The authors would like to thank Michael Egli and Matthias Mueller for their support in the sensor manufacture, and the team at NEURA Robotics for providing the robot arm and product support.


\bibliographystyle{IEEEtran}
\bibliography{references}

\end{document}